\documentclass[11pt]{article}

\usepackage[final]{acl}

\usepackage{times}
\usepackage{latexsym}

\usepackage[T2A,T1]{fontenc}
\usepackage[utf8]{inputenc}
\usepackage[russian,english]{babel}

\usepackage{amsmath}
\usepackage{microtype}
\usepackage{inconsolata}
\usepackage{graphicx}
\usepackage{booktabs}
\PassOptionsToPackage{hyphens}{url}
\usepackage{hyperref}
\usepackage{xurl}

\title{Znaki: A Russian Fingerspelling Recognition Dataset}
\author{
  Petr Surovtsev\textsuperscript{1} \\
  \texttt{pysurovtsev@daimld.tech}
  \And
  Pavel Korotaev\textsuperscript{1} \\
  \texttt{pkorotaevdm@gmail.com}
  \AND
  Alexander Nagaev\textsuperscript{1} \\
  \texttt{aonagaev@daimld.tech}
  \And
  Alexander Kapitanov\textsuperscript{1} \\
  \texttt{aakapitanov@daimld.tech} \\
  \AND
\textsuperscript{1}\textnormal{Department of Artificial Intelligence and Machine Learning Development, DAIMLD}
}

\begin{document}
\maketitle

\begin{abstract}
\label{sec:abstract}

Fingerspelling — the sequential signing of letters that compose proper names, technical terms, and out-of-vocabulary words — is an integral component of any sign language. Despite Russian Sign Language (RSL) being used by more than 150,000 native signers, no public fingerspelling dataset for RSL exists, in contrast to American, British, German, and several other sign languages. We introduce Znaki\footnote{\url{https://github.com/ai-forever/HandReader}}, the first open RSL fingerspelling dataset: 37,252 video recordings (35.4 hours) of 1,593 unique sequences — proper names sampled from 16 domain categories — recorded by 68 native RSL signers (37 Deaf, 29 hard-of-hearing, and 2 certified interpreters) through a crowdsourcing pipeline with a two-stage signer qualification procedure and expert validation. The dataset covers all 33 Cyrillic handshapes, and its letter-frequency distribution matches the Russian National Corpus. As reference points for future research, we establish two baselines adapted from American Sign Language fingerspelling recognition: a Squeezeformer-based encoder-decoder, reaching a character-level accuracy ($1-\mathrm{CER}$) of 0.932 and a sequence accuracy of 0.596, and Fingerspelling-PoseNet, reaching 0.886 and 0.545, respectively.

\end{abstract}

\section{Introduction}
\label{sec:introduction}
Sign languages are full-fledged natural languages that exhibit cultural and regional variation.
There are approximately 72 million Deaf and hard-of-hearing people worldwide, and more than 300 sign languages\footnote{\url{https://education.nationalgeographic.org/resource/sign-language/}}.
Sign language is typically the first (native) language of Deaf individuals, and without adequate education, it may remain their only language — which can hinder communication and limit access to certain products or services.
Given the limited number of sign language interpreters, automatic sign language recognition systems can help reduce communication barriers and contribute to preserving the culture of these languages. As of 2024, there are over 150,000 Deaf signers of Russian Sign Language (RSL) in Russia\footnote{\url{https://voginfo.ru/about/}}. The use of RSL also extends to the countries of the former Soviet Union (now the CIS countries), possibly with minor regional variations, which implies significantly wider use of RSL in the world.

Sign language comprises manual and non-manual components.
Signs can be classified based on their articulation into one-handed and two-handed signs, with two-handed signs being either symmetrical (both hands perform identical movements) or asymmetrical (one hand performs the primary movement while the other serves a supporting role). In RSL, each of the four manual components of a sign — handshape, location, orientation, and movement — contributes to the meaning of the sign.
In RSL, fingerspelling is primarily used to represent proper names, technical and specialized terms, words for which no conventional sign exists, as well as words that are conventionally fingerspelled — such as \foreignlanguage{russian}{``зал''} (\textit{zal}, ``hall'') and \foreignlanguage{russian}{``как''} (\textit{kak}, ``how''). The ability to recognize fingerspelling — an integral part of sign language — is an important step toward full sign language recognition.

\begin{figure}[h]
  \centering
  \includegraphics[width=0.95\linewidth]{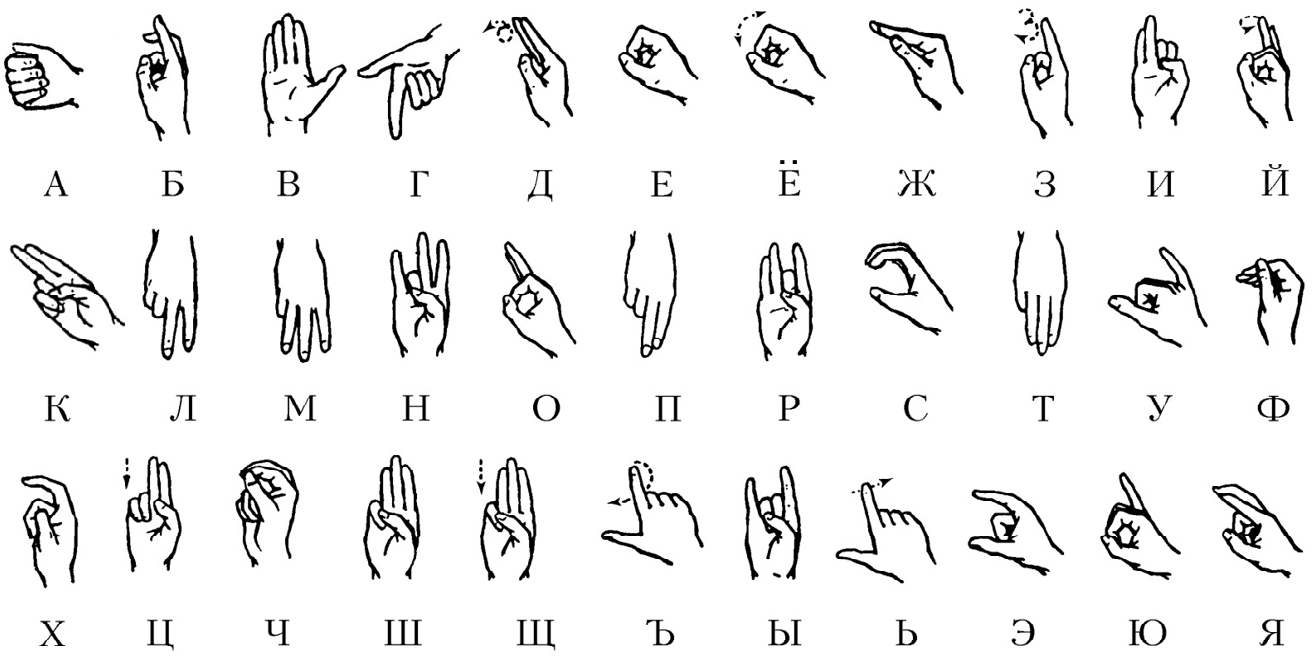}
  \caption{The 33 signs of the RSL alphabet, each corresponding to a letter of the Russian alphabet. Arrows indicate the direction of movement for 8 dynamic signs.}
  \label{fig:alphabet}
\end{figure}

The Russian language has 33 letters, which correspond to 33 unique fingerspelling signs (handshapes) in RSL (see Figure~\ref{fig:alphabet}).
Handshapes are produced with a single dominant hand, approximately at chest level, with each letter being clearly articulated.
Fingerspelling recognition is complicated by the presence of dynamic signs (\foreignlanguage{russian}{``Д'', ``Ё'', ``З'', ``Й'', ``Ц'', ``Щ'', ``Ъ'', ``Ь''} — \textit{D, Yo, Z, Y, Ts, Shch, hard sign, soft sign}) and signs with similar hand configurations — such as \foreignlanguage{russian}{``Ш''}~(\textit{Sh}) and \foreignlanguage{russian}{``Щ''}~(\textit{Shch}), \foreignlanguage{russian}{``Е''}~(\textit{E}) and \foreignlanguage{russian}{``Ё''}~(\textit{Yo}), \foreignlanguage{russian}{``И''}~(\textit{I}) and \foreignlanguage{russian}{``Й''}~(\textit{Y}), and \foreignlanguage{russian}{``Р''}~(\textit{R}) and \foreignlanguage{russian}{``Н''}~(\textit{N}).

This paper presents an efficient data collection methodology and a new fingerspelling dataset for Russian Sign Language recognition.
The main contributions of this paper are as follows:
\begin{itemize}

    \item Znaki, the first publicly available fingerspelling dataset for RSL, comprising 37,252 videos (35.4 hours) of 1,593 unique proper names recorded by 68 native RSL signers.

    \item A Deaf-community-informed phrase sampling covering all 33 Cyrillic handshapes and 696 letter bigrams, with a letter-frequency distribution matching the Russian National Corpus.

    \item A two-stage signer qualification procedure — an RSL proficiency exam followed by a fingerspelling distractor test.

    \item Baseline results from two pose-based fingerspelling recognition architectures previously developed for American Sign Language (ASL).

\end{itemize}

\section{Related Work}
\label{sec:related}

Sign language recognition has been an active area of research for many years. Early approaches primarily focused on image-based recognition of static hand signs in ASL \cite{new2dstatic, realtime}, RSL \cite{makarov}, BdSL \cite{bangla}, and other sign languages \cite{polish, kz}. Nevertheless, dynamic signs are present in virtually all sign languages, and recognizing them necessitates capturing the temporal dimension of the sign.
Robust fingerspelling recognition therefore requires a dataset comprising video samples. At present, two categories of fingerspelling data are publicly available: isolated handshapes (individual letters) and continuous fingerspelling (complete words or phrases). Table~\ref{tab:datasets} summarizes the publicly available resources most relevant to our work.

As noted in Section~\ref{sec:introduction}, no continuous fingerspelling datasets exist for RSL.

\begin{table*}[t]
\centering
\begin{tabular}{llrrrl}
\toprule
\textbf{Dataset} & \textbf{Language} & \textbf{Samples} & \textbf{Signers} & \textbf{Hours} & \textbf{Type} \\
\midrule
RWTH \cite{rwth}                     & DGS & 1,400   & 20  & ---   & isolated \\
NSL23 \cite{nsl23}                   & NSL & 630     & 14  & ---   & isolated \\
Bukva \cite{bukva}                   & RSL & 3,757   & 155 & ---   & isolated \\
NSTU \cite{grif}                     & RSL & 14,978  & 4   & ---   & isolated (private) \\
ChicagoFSWild \cite{fswild}          & ASL & 7,304   & 168 & ---   & continuous \\
ChicagoFSWild+ \cite{fswildplus}     & ASL & 55,232  & 260 & ---   & continuous \\
FSboard \cite{fsboard}               & ASL & 151,000 & 147 & 266   & continuous \\
FS23K \cite{fs23k}                   & BSL & 23,000  & 27  & 10.5  & continuous \\
BANZ-FS \cite{banzfs}                & BANZSL & 35,028 & 116 & ---  & continuous \\
\midrule
\textbf{Znaki (ours)}                & RSL & 37,252  & 68  & 35.4  & continuous \\
\bottomrule
\end{tabular}
\caption{Publicly available fingerspelling datasets. Znaki is the first continuous fingerspelling dataset for RSL.}
\label{tab:datasets}
\end{table*}

\subsection{Isolated Handshape Datasets}

One of the earliest fingerspelling datasets, the RWTH German Fingerspelling Database \cite{rwth}, contains 1,400 video sequences (320$\times$240 and 352$\times$288 resolution, 25 fps) representing 35 DGS signs recorded from 20 participants across two sessions. Each sign was captured simultaneously by two cameras from different viewpoints under varying lighting and background conditions. Participants wore no special clothing and received no training, making the dataset more representative for unconstrained recognition systems.

The Nepali Sign Language dataset (NSL23) \cite{nsl23} is the first dataset for NSL, containing 630 videos (1,205 signs) representing 36 consonants and 13 vowels. Recordings were collected from 14 volunteers (9 beginners, 5 experts) under various conditions (normal lighting, low lighting, prepared/unprepared, and real-world environments).

For RSL handshapes, two datasets are known: the publicly available Bukva dataset \cite{bukva} and the private NSTU dataset \cite{grif}. The Bukva dataset comprises 3,757 videos covering all 33 letters of the Russian alphabet, with an average of 113 videos per class (49 frames per video). Recorded by 155 signers — including Deaf individuals, RSL learners, and teachers — the dataset represents a wide age range (9 to 81 years). The NSTU dataset comprises 13,412 images and videos of 33 Russian fingerspelling signs recorded from four signers (waist-up, right hand at chest level). Due to the limited signer diversity, the authors augmented the dataset with an additional 1,566 samples collected from YouTube and custom recordings to improve real-world performance.

Moreover, general sign language datasets commonly include a collection of handshapes as well \cite{wlasl, slovo}.

\subsection{Continuous Fingerspelling Datasets}
Among the first large-scale ASL fingerspelling collections are the ChicagoFSWild \cite{fswild} and ChicagoFSWild+ \cite{fswildplus} datasets. ChicagoFSWild was collected from online sources (YouTube, ASLized, and deafvideo.tv). The authors collected 214 raw ASL videos and manually annotated all fingerspelling clips, resulting in 7,304 sequences from 168 signers. The dataset includes proper nouns, other words, and abbreviations. Like ChicagoFSWild, the ChicagoFSWild+ dataset was also collected from online sources; however, annotation of the fingerspelling sequences was performed through a crowdsourcing platform. ChicagoFSWild+ contains 55,232 sequences from 260 signers in total, split into 50,402 training sequences (216 signers), 3,115 development sequences (22 signers), and 1,715 test sequences (22 signers).

The FSboard dataset \cite{fsboard} contains 151,000 sequences with an average length of 21 characters, totaling 3.2 million characters and 266 hours of video — recorded entirely on smartphones — making it the largest previously available ASL fingerspelling dataset. It includes 147 unique signers split into training (117), validation (15), and test (15) sets with no phrase overlap between splits. FSboard includes numbers alongside fingerspelling and covers five phrase domains: MacKenzie phrases, random URLs, random street addresses, random names, and random phone numbers.

The FS23K dataset \cite{fs23k} contains 23K BSL fingerspelling instances with letter-level annotations and 133K temporal boundaries. It comprises 23K words, 139K letter annotations, 920K frames, 10.5 hours of video, and 27 signers. The letter distribution closely follows natural frequencies from BBC broadcast data.

The most recent resource, BANZ-FS \cite{banzfs}, is the first large-scale dataset of two-handed fingerspelling in BANZSL — the manual alphabet shared by British, Australian, and New Zealand Sign Languages. It comprises 35,028 fingerspelling instances (over 200K fingerspelled characters) from 116 signers, compiled from three heterogeneous sources: sign-interpreted news broadcasts, controlled laboratory recordings, and web vlogs. Each instance carries multi-level alignment (video to subtitles, fingerspelled letters, and target lexicons), and the dataset supports four benchmark tasks: isolated fingerspelling recognition, fingerspelling detection, detection-then-recognition, and fingerspelling recognition in the context of continuous signing.

Znaki differs from these recent resources in both language and collection methodology. FSboard, FS23K, and BANZ-FS cover ASL and BANZSL — with BANZ-FS specifically targeting the two-handed BANZSL alphabet — whereas Znaki addresses the one-handed manual alphabet of RSL, for which no continuous fingerspelling data was previously available. Furthermore, unlike datasets mined from broadcast or web footage, all Znaki recordings were collected specifically for the dataset through a crowdsourcing pipeline with a two-stage signer qualification procedure and subsequent expert validation, providing explicit control over signer proficiency, phrase coverage, and annotation quality.

\section{Dataset Collection}
\label{sec:dataset}
\begin{figure}[ht]
  \centering
  \includegraphics[width=0.9\linewidth]{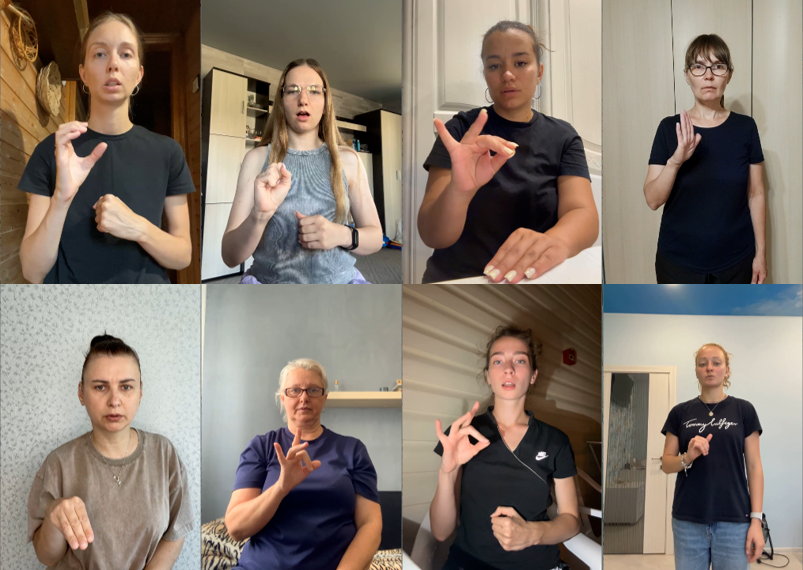}
  \caption{Sample frames from the Znaki dataset.}
  \label{fig:samples}
\end{figure}

Sign language datasets are typically collected using one of three common approaches: (1) studio recordings, (2) recordings sourced from the internet or television broadcasts, and (3) crowdsourcing platforms. The first method suffers from several limitations: a small number of participants and limited clothing diversity; limited variation in backgrounds and lighting; low diversity of recording devices; and overly idealized signing resulting from expert supervision during recording. The limitations of the second approach include a limited number of signers and, most critically, the necessity of additional data filtering, processing, and annotation. The third method avoids many of the limitations listed above; however, it requires a well-established data collection pipeline involving expert validation procedures.
The Znaki dataset was collected using the crowdsourcing platform TagMe\footnote{\url{https://tagme.sberdevices.ru}}. Figure~\ref{fig:samples} provides samples from the Znaki dataset.
The data collection pipeline consisted of the following stages: signer recruitment, video collection, video validation, video processing, and video annotation.

\subsection{RSL Experts}
RSL experts included Deaf or hard-of-hearing individuals and certified interpreters with relevant qualifications. The majority were members of the VOG (All-Russian Society of the Deaf)\footnote{\url{https://voginfo.ru/}} community, including RSL researchers and teachers. These experts were responsible for screening participants during the project admission stages and were also granted access to video verification and annotation tasks.

\subsection{Phrase Selection}
\label{subsec:phrases}
Since fingerspelling is primarily used for signing proper names, only words and phrases belonging to this group were selected for the dataset. We consulted with Deaf experts and identified the 16 most common categories of proper names (see Table~\ref{tabl:categories}), selecting approximately 100 well-known words and phrases for each category. The proper names for each category were generated using GPT-4o \cite{gpt4o}, a large language model (LLM) developed by OpenAI (the prompt template is provided in Appendix~\ref{sec:appendix_prompt}). All generated words and phrases were subsequently verified by RSL experts and manually refined when necessary. The resulting phrases contain between 1 and 4 proper names, with an average of 1.4 words (\(\sigma = 0.6\)), and range from 3 to 29 (excluding the space character) characters in length, with an average of 10,0 characters (\(\sigma = 4.1\)).

\begin{table}[ht]
\centering
\scalebox{0.95}{
\begin{tabular}{lc}
\hline
\textbf{Category} & \textbf{Count} \\
\hline\hline
Moscow Metro                  & 231 \\
Social terms                  & 121 \\
Banking terms                 & 107 \\
Countries                     & 99  \\
Russian cities                & 99  \\
Authors                       & 98  \\
Book titles                   & 97  \\
Government institutions       & 96  \\
World cities                  & 93  \\
Movie titles                  & 92  \\
World landmarks               & 91  \\
Disease names                 & 90  \\
Russian landmarks             & 84  \\
Professions                   & 75  \\
Geographical names            & 70  \\
Personal names                & 50  \\
\hline
\end{tabular}}
\caption{The number of unique samples in each category of the Znaki dataset.}
\label{tabl:categories}
\end{table}

\subsection{Signer Selection}
\label{subsec:signers}
We specifically sought to recruit Deaf or hard-of-hearing signers who are members of the VOG community. However, the platform did not restrict projects to specific user groups — they were accessible to all annotators. To ensure data quality, we implemented a two-stage screening process.

\vspace{0.5em} \noindent\textbf{RSL knowledge exam.} Candidates were required to pass a knowledge test of RSL, which involved recording a video translating Russian sentences into RSL. RSL experts and RSL teachers evaluated the correctness of the submissions and determined whether candidates would proceed to the next test. This exam made it possible to assess the candidates' RSL proficiency and their ability to understand and follow task instructions.

\vspace{0.5em} \noindent\textbf{Fingerspelling exam.} Candidates who passed the RSL exam were then required to complete an additional fingerspelling test. This test consisted of 15 tasks. In each task, candidates watched a video of RSL fingerspelling and had to select the correct Russian translation from a list of words, with a time limit of 30 seconds per task. The videos were recorded by RSL experts, who also created the incorrect response options. These experts deliberately designed plausible distractors that could be mistaken for the correct answer due to a candidate's lack of knowledge or attention. A score of at least 85\% correct answers was required to pass. 68 candidates out of 100 were accepted to the main tasks.

\subsection{Video Collection}
\label{subsec:video_collection}
The video collection task was accompanied by written instructions describing the recording rules and quality requirements.
Additionally, signers were provided with a video in which RSL teachers explained the same instructions in sign language.

The recording rules were as follows: (1) only one person could appear in the video; (2) the signing hand had to remain fully within the frame; and (3) camera shaking and visual effects were not permitted.
The platform verification process accepted only videos with a minimum HD resolution, a duration exceeding one second, and a frame rate of at least 15 fps.
Duplicate videos were rejected, and signers were allowed to re-record submissions.

\begin{figure}[h]
  \centering
  \includegraphics[width=0.6\linewidth]{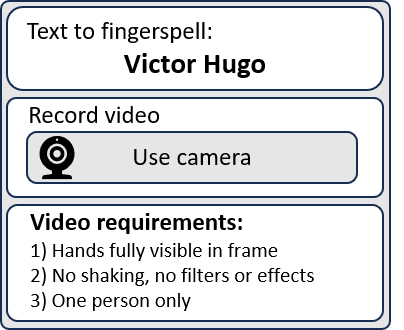}
  \caption{Task for recording the proper name \foreignlanguage{russian}{«Виктор Гюго»} (Victor Hugo).}
  \label{fig:mining}
\end{figure}

Signers could record videos in arbitrary environments, making the dataset heterogeneous with respect to recording conditions.
To further increase data diversity, a daily limit of 30 tasks per signer was enforced.
Figure~\ref{fig:mining} illustrates an example of a video recording task.

\subsection{Video Validation}
\label{subsec:video_validation}
Each video was checked for the correctness of the translation from Russian into RSL fingerspelling sequences.
The videos were validated by a diverse group of participants, including Deaf and hard-of-hearing individuals, certified RSL interpreters, and RSL teachers and educators, all of whom confirmed their proficiency in Russian Sign Language. Validators also verified that each video complied with the recording requirements.

To become a validator, crowdworkers were required to complete the mandatory qualifying tasks described in Section~\ref{subsec:signers}, undergo training, and pass a video validation exam.

Each video was reviewed by at least three validators.
In cases of insufficient inter-annotator agreement, additional validators were recruited until a satisfactory level of consensus was achieved. Overall, 82\% of the samples received unanimous validator agreement, i.e., inter-annotator agreement of the validation judgments, indicating a high level of annotation reliability. On average, validating a single video took 17.3 seconds.

Approximately 12\% of the collected videos were rejected during expert validation due to signing errors or failure to comply with the recording requirements and instructions.

\subsection{Time Interval Annotation}
\label{subsec:video_annotation}
\begin{figure}[ht]
  \centering
  \includegraphics[width=0.99\linewidth]{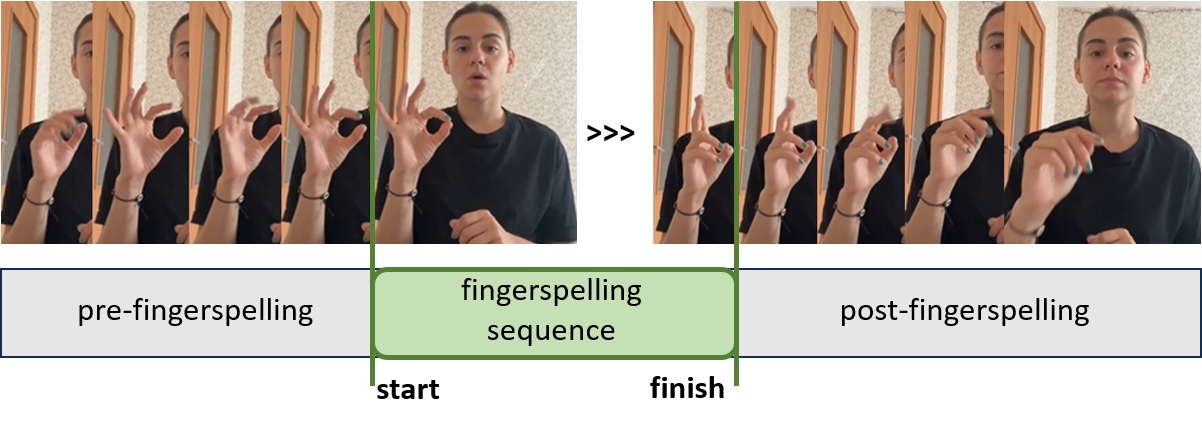}
  \caption{Illustration of the annotation scheme for fingerspelling sequence boundaries (start and end).}
  \label{fig:marks}
\end{figure}

Time interval annotation was required to separate the sign sequence from other activity periods (see Figure~\ref{fig:marks}). As in the validation stage, crowdworkers received training and passed an exam to gain access to the annotation tasks.
Validated videos were processed to 30 frames per second (fps) to ensure consistent time alignment. Each video was annotated by three users, and the annotations were subsequently aggregated into a single result using an open-source aggregation framework\footnote{\url{https://github.com/ai-forever/aggme}}; 95\% of the videos achieved an agreement level above 80\%. This annotation approach made it possible to determine the fingerspelling speed, with an average rate of 3.2 signs (letters) per second. Importantly, the signing rate was not strongly dependent on the sequence length, as the standard deviation was only 0.1 signs per second.

Note that the annotations mark the boundaries of the entire fingerspelled sequence (start and end); the dataset does not include frame-level or letter-level segmentation. Reliable letter boundaries are difficult and costly to annotate at scale, since adjacent handshapes blend into one another due to coarticulation (cf. FS23K \cite{fs23k}, where letter boundaries were annotated for a substantially smaller collection). We discuss the implications of this design choice in the Limitations section.

\subsection{Dataset Characteristics}

\begin{figure}[ht]
  \centering
  \includegraphics[width=0.9\linewidth]{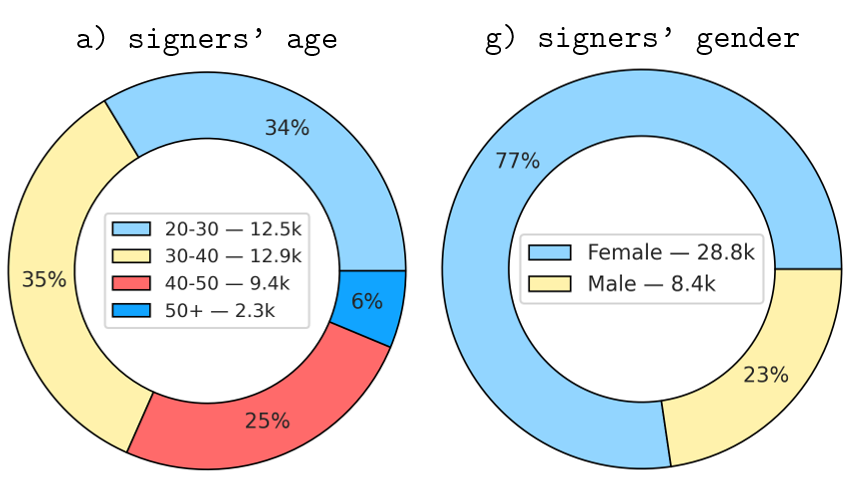}
  \caption{Age and gender distribution in the Znaki dataset.}
  \label{fig:age_gender}
\end{figure}

\vspace{0.5em} \noindent\textbf{Signers.}
The dataset was recorded by 68 signers, including 37 Deaf individuals, 29 hard-of-hearing individuals, and two sign language interpreters (see Figure~\ref{fig:age_gender}). The signers are aged from 20 to 54 years, with 52 women and 16 men.

\begin{figure}[ht]
  \centering
  \includegraphics[width=0.95\linewidth]{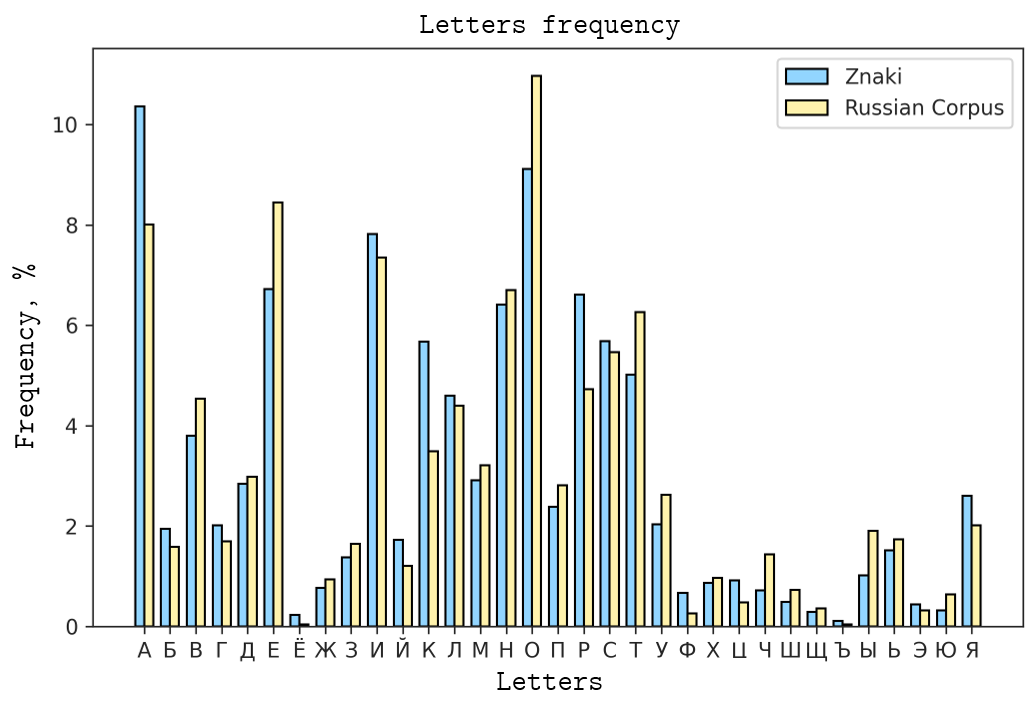}
  \caption{Comparison of letter frequencies in the Znaki dataset and the Russian National Corpus.}
  \label{fig:letts_freq}
\end{figure}

\vspace{0.5em} \noindent\textbf{Sequences.}
The dataset contains 1,593 unique proper names, covering all 33 letters of the Russian alphabet and 696 letter bigrams ($\approx$64\% of all possible combinations). The letter-frequency distribution in Znaki corresponds to that of the Russian National Corpus\footnote{\url{https://ruscorpora.ru/en/}} (see Figure~\ref{fig:letts_freq}).

\begin{figure}[ht]
    \centering
    \includegraphics[width=0.85\linewidth]{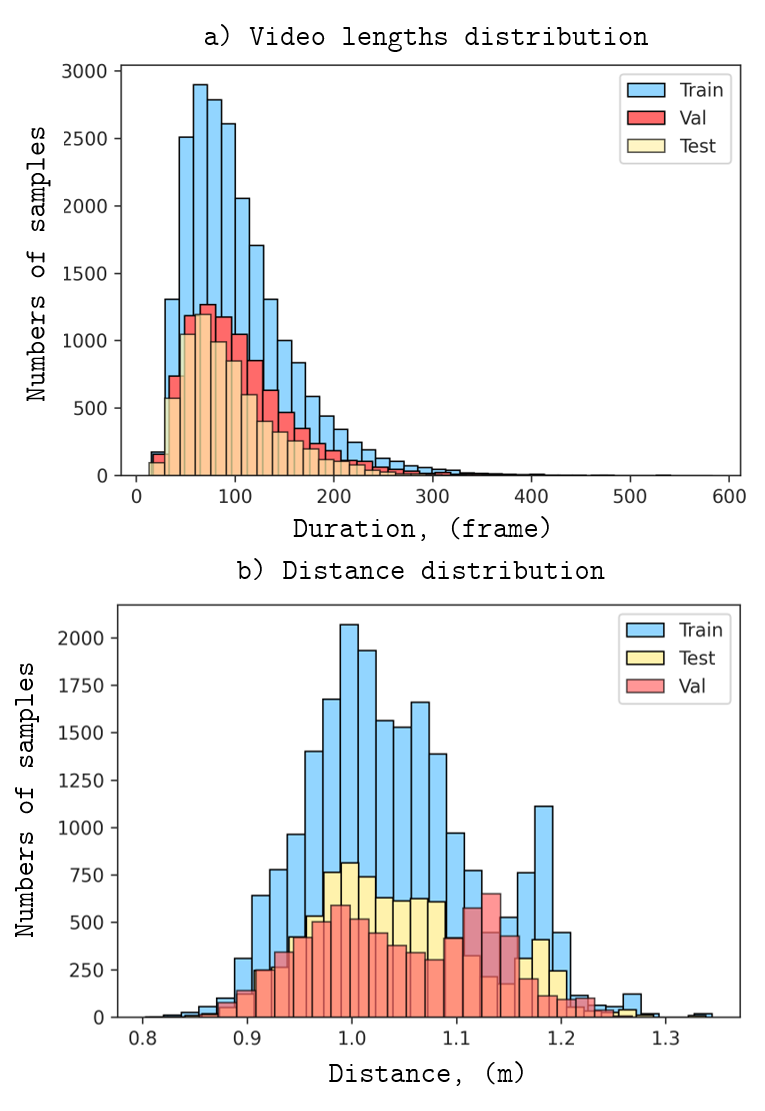}
    \caption{Znaki dataset statistics: a) video length distribution; b) distance distribution, where the distance to the camera is approximately estimated in meters by computing the length between the signer's left and right shoulders using MediaPipe \cite{mediapipe}.}
    \label{fig:statistics_dataset}
\end{figure}

\vspace{0.5em} \noindent\textbf{Videos.}
Znaki contains 37,252 videos (35.4 hours in total) with a minimum frame side of 720 pixels, trimmed according to the time interval annotations, resulting in about 42~GB of data.
Each sequence was recorded by between 8 and 41 different signers, averaging 23 videos per phrase. The average length of each sequence is 102 frames (see Figure~\ref{fig:statistics_dataset}). More than 80\% of the dataset's videos are in Full HD format, and the rest are in HD. About 88\% of the samples were recorded on a phone camera, and the rest on a laptop.

\begin{figure}[!ht]
    \centering
    \includegraphics[width=0.95\linewidth]{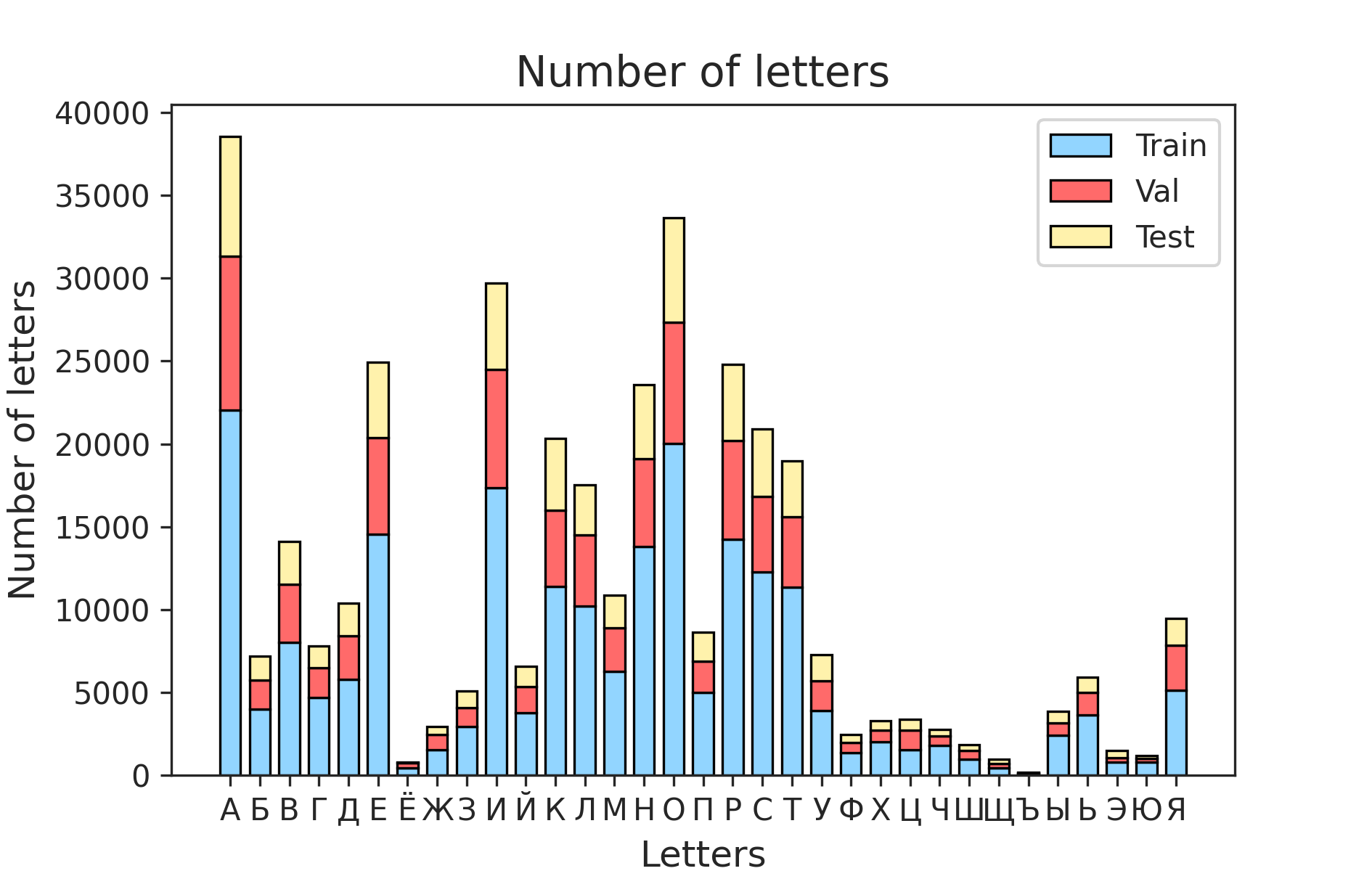}
    \caption{Distribution of the total number of letters across the dataset splits.}
    \label{fig:num_letters}
\end{figure}

\vspace{0.5em} \noindent\textbf{Split.}
The initial collection comprised more than 50,000 videos. To avoid potential data leakage — specifically, to ensure no overlap of phrases or signers between training, validation, and test sets — a portion of the data was deliberately excluded from the final dataset. We split the dataset into training (58\%), validation (23\%), and test (19\%) sets, each containing unique phrases only: 743, 300, and 550, respectively. No signers from the test set overlap with the training and validation sets. Additionally, we ensured a proportional distribution of letters across the splits (see Figure~\ref{fig:num_letters}).

\section{Experiments}
\label{sec:experiments}

\subsection{Experimental Setup}
The goal of our experiments is not to introduce a new recognition architecture, but to validate the proposed dataset as a viable benchmark for RSL fingerspelling recognition. Since no publicly available benchmark for this task currently exists, we establish baseline results using two publicly available fingerspelling recognition models \cite{christofhenkel, fsp_posenet} originally developed for ASL.

All experiments were conducted on a single H100 80~GB GPU with 12 CPU cores and 244~GB RAM. Training each model required approximately 14 GPU hours. Training hyperparameters are listed in Appendix~\ref{sec:appendix_hyper}.

\vspace{0.5em} \noindent\textbf{Dataset Split.}
No additional preprocessing or modifications were applied to the proposed train/validation/test split. The models were trained on the training set, validated on the validation set, and evaluated on the test set.

\vspace{0.5em} \noindent\textbf{Input Representation.}
We used MediaPipe Holistic \cite{grishchenko2020mediapipe} to extract a total of 543 landmarks per frame: 33 pose landmarks, 21 landmarks for each hand, and 468 facial landmarks.

\vspace{0.5em} \noindent\textbf{Evaluation Metrics.}
We evaluate our models using two complementary metrics: Character Error Rate (CER) and Accuracy (Acc). CER measures the edit distance between the predicted and ground-truth character sequences. To improve interpretability (higher is better), we report \(1 - \mathrm{CER}\), where a value of 1.0 corresponds to perfect character-level prediction:

\[
1 - \mathrm{CER} = 1 - \frac{S + D + I}{N},
\]

where \(S\) is the number of substitutions, \(D\) is the number of deletions, \(I\) is the number of insertions, and \(N\) is the total number of characters in the ground-truth sequence.

Accuracy is defined as the proportion of completely correct sequence predictions (all characters in the correct order) relative to the total number of test samples:

\[
\mathrm{Acc} = \frac{\text{\# fully correct sequences}}
{\text{\# total sequences}}.
\]

\subsection{Baseline Models}
We benchmark the dataset using two publicly available architectures previously proposed for ASL fingerspelling recognition.

\vspace{0.5em} \noindent\textbf{Transformer-Based.} Our first baseline is based on the winning solution of the Kaggle ASL Fingerspelling competition \cite{christofhenkel}. The architecture employs an improved Squeezeformer encoder adapted for MediaPipe landmark sequences together with a two-layer transformer decoder for sequence prediction. From the full set of extracted landmarks, the model uses a subset of 130 per frame: 42 hand landmarks (21 per hand), 12 pose landmarks, and 76 facial landmarks.

\vspace{0.5em} \noindent\textbf{Fingerspelling-PoseNet.} Our second baseline is based on Fingerspelling-PoseNet \cite{fsp_posenet}, a pose-based transformer encoder-decoder architecture designed for fingerspelling recognition from hand pose sequences. The model operates on the 21 2D landmarks of the automatically detected signing hand and performs autoregressive sequence prediction.

\vspace{0.5em} \noindent\textbf{Adaptation to RSL.} Both baselines were transferred to RSL with minimal changes. The only modification was a configuration-level replacement of the output vocabulary with the 33 letters of the Cyrillic alphabet; the source code of both public implementations was left unchanged, so the input representation, normalization, architecture, optimizer settings, learning rate schedule, and augmentations remain the defaults released by the original authors (see Appendix~\ref{sec:appendix_hyper}).

\subsection{Results}
We evaluate the two baseline architectures on the held-out test split of the Znaki dataset. Table~\ref{tab:baseline_results} reports character-level \(1 - \mathrm{CER}\) and phrase-level (Acc) performance.

\begin{table}[ht]
\centering
\begin{tabular}{lcc}
\hline
Model & \(1 - \mathrm{CER}\) $\uparrow$ & Acc $\uparrow$ \\
\hline
FS-PoseNet & 0.886 & 0.545 \\
Transformer-Based & \textbf{0.932} & \textbf{0.596} \\
\hline
\end{tabular}
\caption{Baseline results on the proposed RSL fingerspelling dataset.}
\label{tab:baseline_results}
\end{table}

The Transformer-Based model, adapted from ASL to RSL, achieves strong character-level accuracy (\(1-\mathrm{CER} = 0.932\)) but exhibits a substantial gap at the sequence level (Acc = 0.596), indicating compounding errors in sequence prediction.

Error analysis reveals three primary failure modes:
\begin{itemize}
    \item \textbf{Visual confusability}: high confusion rates among visually similar handshapes (e.g., \foreignlanguage{russian}{``Ш''/``Щ''} (\textit{Sh/Shch}), \foreignlanguage{russian}{``Е''/``Ё''} (\textit{E/Yo}), \foreignlanguage{russian}{``И''/``Й''} (\textit{I/Y}), \foreignlanguage{russian}{``Р''/``Н''} (\textit{R/N})), accounting for 38\% of character substitutions.
    \item \textbf{Dynamic sign instability}: temporal segmentation errors for movement-based handshapes (\foreignlanguage{russian}{``Д'', ``З'', ``Й'', ``Ц'', ``Щ''} — \textit{D, Z, Y, Ts, Shch}), leading to 27\% of deletions and insertions.
    \item \textbf{Length-dependent degradation}: Acc decreases by 4.2\% per additional character for sequences of $\geq$4 letters, reflecting accumulated alignment uncertainty.
\end{itemize}

Despite heterogeneous recording conditions (lighting, viewpoint, device), models trained on MediaPipe Holistic landmarks \cite{grishchenko2020mediapipe} demonstrate robust generalization: no performance drop is observed when evaluating on signers unseen during training, confirming the absence of signer-specific overfitting.

These baselines establish Znaki as a challenging benchmark for RSL fingerspelling recognition and provide a reference point for future methodological advances.

\section{Conclusion}
\label{sec:conclusion}
In this work, we presented Znaki — the first public fingerspelling dataset for RSL and the largest RSL fingerspelling resource to date. By involving only native and near-native members of the signing community, Znaki captures natural intra- and inter-signer variability while ensuring the linguistic authenticity of the collected data, addressing a critical gap: prior to our work, no publicly available RSL fingerspelling datasets existed.

We involved RSL experts in defining the categories of lexical sequences and used LLMs to generate candidate proper names for recording. In addition, we developed a participant verification procedure for the crowdsourcing platform to improve recording quality and annotation reliability.

To validate the usability of the proposed dataset, we trained two publicly available strong baselines previously developed for ASL fingerspelling recognition. The successful transfer of these architectures to RSL fingerspelling demonstrates the suitability of the proposed dataset for training and evaluating modern fingerspelling recognition models.

We believe that publicly releasing Znaki will stimulate further progress in sign language processing, particularly for under-resourced sign languages.

\section*{Limitations}
\label{sec:limitations}

The Znaki dataset reflects the demographic composition of its participants and therefore exhibits several imbalances. The dataset was collected from signers aged 20--54 years, without representation of children or elderly signers, whose signing patterns may differ substantially. In addition, the gender distribution is imbalanced (52 women and 16 men), and all participants belong to a relatively homogeneous ethnic and geographic background. These factors may limit the generalization of models trained on Znaki to broader signer populations.

The dataset focuses exclusively on fingerspelled proper names and predefined phrase categories. As a result, it does not cover several important aspects of real-world fingerspelling communication, including numerical sequences, conversational phrases, abbreviations, partial fingerspelling, fingerspelling embedded in mixed lexical contexts (interleaved with lexical signs), and coarticulation between fingerspelling and surrounding signs in continuous signing. Although the selected phrase categories were designed in consultation with Deaf experts, the phrase distribution may not fully reflect natural fingerspelling usage frequencies in the Deaf community. More broadly, this scope is deliberate: Znaki is a fingerspelling dataset rather than a full continuous sign language recognition or translation resource, and it does not include RSL gloss annotations — fingerspelling is one component of sign language communication, not a substitute for full sign language recognition.

The annotations provide only sequence-level temporal boundaries (start and end of fingerspelling); frame-level and letter-level segmentation is not available. This limits the direct use of Znaki for research on letter-to-frame alignment, fine-grained temporal modeling of handshape transitions, and fingerspelling detection within continuous signing. Adding letter-level temporal annotations is a natural direction for future extensions of the dataset.

The recording setup also introduces several constraints. Most videos were recorded using smartphone cameras in frontal-view settings, with limited variation in viewpoint. The dataset does not explicitly cover challenging scenarios such as severe occlusions, extreme lighting conditions, or multi-view recordings.

Our experimental evaluation is limited to two publicly available baseline architectures originally developed for ASL fingerspelling recognition. Therefore, the reported results should be viewed as initial baselines rather than exhaustive performance estimates for the dataset.

\section*{Ethical Statement}
\label{sec:ethical}

The legal and ethical aspects of the dataset collection procedure were reviewed and approved by our institution's legal team. All crowdworkers provided informed consent through the crowdsourcing platform, authorizing the processing and public release of their data for research purposes. Participation in the project was voluntary, and all contributors received financial compensation proportional to the time required for task completion. Compensation rates were aligned with the average income levels in the Russian Federation.

To protect participant privacy, all contributors were identified only through anonymized hash-based user identifiers. Since facial expressions and non-manual components constitute an essential part of sign language communication, facial information was retained in the recordings and may be misused in surveillance-related applications.

To mitigate such risks, the dataset is released exclusively for research purposes under a restricted-use license.

\bibliography{main}

\appendix

\section{Phrase Generation Prompt}
\label{sec:appendix_prompt}

Candidate proper names for each of the 16 categories (Table~\ref{tabl:categories}) were generated with GPT-4o using the following prompt template:

\begin{quote}
\textit{Compile a list of the 100 most popular proper names from the ``\{category\}'' domain.}
\end{quote}

All generated candidates were subsequently verified by RSL experts and manually refined when necessary (see Section~\ref{subsec:phrases}).

\section{Training Details}
\label{sec:appendix_hyper}

Both baselines were trained on the training split with early stopping on the validation split, using the default hyperparameters of the respective public implementations \cite{christofhenkel, fsp_posenet}. We did not modify the source code of either repository: the training procedure, optimizer settings, learning rate schedule, and augmentations were kept exactly as released by the original authors, so the exact values can be found in the corresponding public repositories. The only adaptation required for RSL was a configuration-level change of the output vocabulary, replacing the original character set with the 33 letters of the Cyrillic alphabet. Each model was trained for approximately 14 hours on a single H100 80~GB GPU.

\end{document}